# IfcWoD,Semantically Adapting IFC Model Relations into OWL Properties

Tarcisio Mendes de Farias, t.mendesdefarias@active3d.net
*ACTIVe3D, Dijon, France*

Ana-Maria Roxin, ana-maria.roxin@u-bourgogne.fr
*Checksem, LE2I UMR6306, CNRS, ENSAM, Univ. Bourgogne Franche-Comté, F-21000 Dijon, France*

Christophe Nicolle, cnicolle@u-bourgogne.fr
*Checksem, LE2I UMR6306, CNRS, ENSAM, Univ. Bourgogne Franche-Comté, F-21000 Dijon, France*

## Abstract
In the context of Building Information Modelling, ontologies have been identified as interesting in achieving information interoperability. Regarding the construction and facility management domains, several IFC (Industry Foundation Classes) based ontologies have been developed, such as IfcOWL. In the context of ontology modelling, the constraint of optimizing the size of IFC STEP-based files can be leveraged. In this paper, we propose an adaptation of the IFC model into OWL which leverages from all modelling constraints required by theobject-oriented structure of the IFC schema. Therefore, we do not only present a syntactic but also a semantic adaptation of the IFC model. Our model takes into consideration the meaning of entities, relationships, properties and attributes defined by the IFC standard. Our approach presents several advantages compared to other initiatives such asthe optimization of query execution time. Every advantage is defended by means of practical examples and benchmarks.

Keywords:IFC, Linked Data, OWL, Ontology, Building Information Modelling (BIM)

## 1 Introduction
Linked Data has been recognized as a promising research field for publishing and interlinking heterogeneous data from different Web repositories(Nešić, et al., 2011). For doing so, Linked Data relies on four main principles(Bizer, et al., 2009), derived from traditional Web architectures' principles(Heath & Bizer, 2011). First, resources are uniquely and globally identified by means of URIs (Uniform Resource Identifiers). Second, resources have to be accessed using the HTTP protocol, therefore HTTP URIs have to be used. Third, resource descriptions have to be delivered using the RDF standard model; resource queries have to be specified in SPARQL standard language. Finally, resource descriptions have to include RDF links to other resources (from other datasets), in order to increase the possibilities for discovering new resources.

In the context of BIM (Building Information Modelling), such modelling of resourceshas been identified as interesting approach for achieving information interoperability(Pauwels, et al., 2011)(Farias, et al., 2014). Indeed, today's ISO standard for BIM information exchange is the IFC (Industry Foundation Classes) model(ISO , 2013).IFC was designed in order to optimize the size of STEP-based files(ISO, 2002)exchanged among project stakeholders. Still, in the context of ontology modelling, constraints imposed by IFC object-oriented modelling principles can be leveraged.  Moreover, by implementing semantic adaptations of the IFC standard one can implement more intuitive building information manipulation. This is one of the reasons why, in construction and facility management domains, severalIFC-based ontologies have been developed.Existing IFC-related ontologies were conceived as direct syntax mappings between EXPRESS and OWL languages(Beetz, et al., 2009)(Pauwels, et al., 2011). One of the latest and the most solid implementationsof an IFC ontology is IfcOWL proposed in(W3C Linked Building Data Community Group, 2014).

In this paper, we propose an adaptation of the IFC model into the OWL language. Our model takes advantages of RDFgraph-based structure andOWL's features for reasoning capabilities (data inference), thus leveraging from all modelling constraints required by the object-oriented structure of the EXPRESS language. Therefore, we do not only present a syntactic but also a semantic adaptation of the IFC model. Our model takes into consideration the meaning of the entities, relationships, properties and attributes as defined by the IFC standard. Our approach presents several advantages compared to other initiatives: it simplifies and eases query writing, it optimizes query execution, it maximizes inference capabilities as performed by the reasoners and finally allows reducing data redundancy.

This paper begins with a brief overview of the main related works. Then, section 3 presents our approach. Allprevious stated advantages are defended by means of practical examples and benchmarks in section 4.Finally, we conclude this article in section 5.

## 2 Related Works

Developed in the context of the IntelliGridEU FP6 project(Gehre, et al., 2006), the approach described by Beetz et al. in (Beetz, et al., 2009)is one of the most used approaches for translating the IFC standard into OWL language (IfcOWL). Authors present a semi-automatic method forconceivinganOWL ontology from the EXPRESS schema of the IFC standard. They establish various mapping rules such as every EXPRESS entity is mapped into an OWL class.

Pauwels et al. in (Pauwels, et al., 2011)developed a similar method to IfcOWL for conceiving an IFC OWL ontology, butthey further focused on Linked Open Data aspects(Heath & Bizer, 2011). They also provide a web service for automatically converting the IFC model into itsSemantic Web version.Beetz et al. in (Beetz, et al., 2015)propose anintermediate approach mixing ISO 10303 part 21 STEP Physical File (SPF) format and RDF vocabulary. For doing so, authors map only the *IfcPropertySingleValue* entity into an RDF property. Theyjustified this approach by the fact that IFC geometric representations cannot be as efficiently stored in RDF as in SPF. Nevertheless, for mitigating this problem,in (Farias, et al., 2014), we have proposed to translate EXPRESS collections (e.g. LIST)either into non-functional OWL object properties or multiple objet and data properties,instead of using RDF List or OWL List constructs(Drummond, et al., 2006). We argue thiswith the fact that in cases where the property order is not important, acollection (LIST) can be mapped as several distinct values of a non-functional OWL property. Otherwise,for a finite collection, we can create different data or object properties.For exemplifying this, we may consider the IFC attribute "coordinates" from *IfcCartesianPoint*entity. This attribute contains an ordered list of three elements. With our approach, such attribute valueis mapped tothree OWL object properties: *coordinateX*, *coordinateY* and *coordinateZ*. Doing so allows us, for instance,to directly query the "Z" coordinate of a Cartesian point, without having to parse the entire coordinates list.

Beetz et al. in (Beetz, et al., 2015)proposed a solutionas a first attempt to interpret IFC relation entities as OWL properties. However, their approach is limited to RDF vocabulary and only considers the *IfcPropertySingleValue*entity for property mapping. Furthermore, no method is proposed for semi-automatically or automatically translating *IfcRelationship* and *IfcProperty*(a super-type of *IfcPropertySingleValue*) entities as OWL properties.

In December 2014, the W3C Linked Building Data Community Group released a building ontology (W3C Linked Building Data Community Group, 2014)based on the latest version of the IFC2x4 standard. As it is an approach widely accepted by the Linked Building Data community,it can beconsidered as a state-of-the-art IfcOWL ontology. Still, this ontology was built starting from the EXPRESS specification of the IFC2x4 standard and using a direct mapping of EXPRESS entities to OWL classes, as was the caseinthe previous cited

approaches.This version of IfcOWLalso needs some improvements such as avoiding the usage ofOWL list structures for representing EXPRESS collections (e.g.: LIST or SET) as earlier described. For nextsections of this paper, the term IfcOWLmakes reference to the IFC-based ontology proposed by the W3C Linked Building Data Community(W3C Linked Building Data Community Group, 2014).

## 3 IfcWoD, the IFC Web of DataOntology

The latest version of IfcOWLdoes not allow fully exploiting OWL features, and still suffers from the limitations imposed by the EXPRESS language(STEP Tools, 2015). Indeed, the IFC standard was conceived for supporting object-oriented databases(Lee, et al., 2014). Moreover, one of the main goals of IFC STEP serialization was to optimize the size of exchanged data files. For example, various properties of the IFC entities are indeed a set of *IfcPropertySingleValue*instances which are encapsulated in an *IfcPropertySet* entity. Figure 1illustrates a portion of such IFC-STEP file where several objects of *IfcWallStandardCase*type (e.g.: #3060) point to the same set of *IfcPropertySingleValue*instances (e.g.: #2935, #2936, #2937, #2941). This set of properties is related to *IfcWallStandardCase*objectsthrough one instance of an*IfcRelDefinesbyProperties*entity(e.g.: #14997). This referencing process allows considerably reducing the size of such STEP-based file.This applies to the context of one single STEP file, and is possible since various IFC objects can reference identical relationships only by using the unique STEP entity instance identifier (e.g. in the form “#123”). However, when conceiving the IFC standard ontology, from a semantic point of view, these relationship and property entities would be better “translated” into OWL properties instead of OWL classes and their instances (the considered IfcOWL ontology applies the latter). Doing this does not imply increasing data redundancy (as was the case for STEP-based files), notably because OWL knowledge bases (meaning TBoxandABox) are stored as triples in triple stores (i.e.: semantic graph knowledge bases or RDF stores)(Allemang & A. Hendler, 2008). We further explain this statement below.

```
#2935=IFCPROPERTYSINGLEVALUE('Retournement aux ouvertures',$, IFCINTEGER(0),$);
#2936=IFCPROPERTYSINGLEVALUE('Retournement aux extr\X\E9mit\X\E9s', $,IFCINTEGER(0),$);
#2937=IFCPROPERTYSINGLEVALUE('Largeur',$,IFCLENGTHMEASURE(0.32),$);
#2941=IFCPROPERTYSINGLEVALUE('Fonction',$,IFCINTEGER(0),$);
...
#2953=IFCMATERIAL('Mur par d\X\E9faut');
#2960=IFCMATERIALLAYER(#2953,0.32,$);
#2961=IFCMATERIALLAYERSET((#2960),'Mur de base:G\X\E9n\X\E9rique - 320 mm');
...
#2950=IFCPROPERTYSET('1Ks371pQHCOx_oJSZW78op',#34,'PSet_Revit_Type_Construction',
$,(#2935,#2936,#2937,#2941));
...
#3060=IFCWALLSTANDARDCASE('1iSKq$8HT2UvXyfHrxgRuh',#34,
'Mur de base:G\X\E9n\X\E9rique - 320 mm:193133',$,
'Mur de base:G\X\E9n\X\E9rique - 320 mm:168423',#3036,#3059,'193133');
...
#3085=IFCMATERIALLAYERSETUSAGE(#2961,.AXIS2.,.NEGATIVE.,0.16);
...
#14997=IFCRELDEFINESBYPROPERTIES('3NW$BjkfD1gh$ao1JHSikk',#34,$,$,
(#2890,#2906,#3002,#3060,#4605,#4685,#12594,#12656),#2950);
...
#14895=IFCRELASSOCIATESMATERIAL('0$aut9s4991QbgrydDAIH3',#34,
$,$,(#3060),#3085);
```

**Figure 1**A portion of the IFC-STEP file that exemplifies an IfcRelDefinesByProperties relationship between IfcWallStandardCase objects and a set of IfcPropertySingleValue properties.

Figure 1 shows the definition of several STEP identifiers each one referencing only one instance of an IFC entity. For example, one instance of *IfcRelDefinesbyProperties*(i.e.: #14997) links a set of IFC objects (i.e.: #2890,#2906,#3002,#3060,#4605,#4685,#12594,#12656) to one *IfcPropertySet* instance (i.e.: #2950). The STEP syntax illustrated above allows optimizing the resulting file size. Indeed, it is not necessary to re-write an *IfcRelDefinesbyProperties* relationship for each IFC object implementing the same property set. Without the possibility of specifying such sets as a value for STEP entities' attributes, eight *IfcRelDefinesbyProperties* data entries would be necessary. Let us now suppose the case of the IfcOWL[1] ontologypopulated with data from the IFC file illustrated in Figure 1.We use the prefix *ifcowl* for denoting terms issued from this ontology. Only one instance of type *ifcowl*:*IfcRelDefinesbyProperties*is created for representing data from one *IfcRelDefinesbyProperties*data entry in the STEP format(STEP identifier "#14997" in Figure 1). Moreover, the previous set of IFC objects is interpreted as values for the*ifcowl:RelatedObjects_of_IfcRelDefinesbyProperties*non-functional OWL object property. When stored in a RDF triple store, each property assertion becomes a triple (e.g.::*IfcRelDefinesbyProperties*_14997*ifcowl:RelatedObjects_of_IfcRelDefinesbyProperties: IfcWallStandardCase_3060*).

When considering the definition of an ontology (Guarino, et al., 2009), such data structuring isn't advised as it presents several drawbacks. Indeed, the considered IfcOWL ontology hardensthe understanding of IFC object properties and relationships. Also it complicates the correctapplication of Linked Data principles (notably the implementation ofRDF links to other knowledge bases). Moreover, the TBox model defined for the considered IfcOWL ontology makes query writing very difficult. We explain how our approach can leverage these points in section 4.

Studer et al. in (Studer, et al., 1998)define an ontology merging Gruber's(Gruber, 1993)and Borst's(Borst, 1997)ontology definitions as follows: "An ontology is a formal, explicit specification of a shared conceptualization". Based on this definition, we propose defining IFC properties and relationships in the ontology schema instead of using instances of OWL classes (e.g. *ifcowl:IfcProperty* or *ifcowl:IfcRelationship*). By doing so, our resulting ontologycorrespondsto the above mentioned ontology definition (e.g.: explicit specification). We use the prefix *ifcwod*for denoting terms from our IFC-based ontology (IfcWoD for IFC Web of Data). The methodology used forthe semi-automatic definition ofthe IfcWoD ontology is further described in the next subsections.

### 3.1 Adapting IfcRelationship entity into OWL ontology

Figure 2 exemplifies our approach for modelling IFC relationships (i.e.: subtypes of *IfcRelationship*) semantically adapting into OWL ontologies based on Studer et al. ontology definition. Considering our previous example, we explicitly specify the relation between *ifcowl:IfcWallStandardCase*instances and their property set (*ifcowl*:*IfcPropertySet*instance*)*by asserting OWL properties*ifcwod:isDefinedBy_IfcObject*(e.g. *:IfcWallStandardCase_3060ifcwod:isDefinedBy_IfcObjectIfcPropertySet_2950*).In our approach, we consider this property to be the representation of the semantic meaning of the IFC entity *IfcRelDefinesbyProperties.*

In the IFC model, the*IfcRelationship*entity is a super-type (or a superclass in the IfcOWL ontology)of several relationships possible among IFC objects. These relationships represent the major building'ssemantics. Still, in IfcOWL, these relations are stated as OWL classes. Hence, a relation is expressed in the ontology'sABox as an instance of a subclass of *IfcRelationship*. Although, a relation can be explicitly defined as an OWL object property, these relations in IfcOWLare not explicitly specified in the schema (i.e.: TBox).

[1] http://linkedbuildingdata.net/resources/IFC4_ADD1.owl

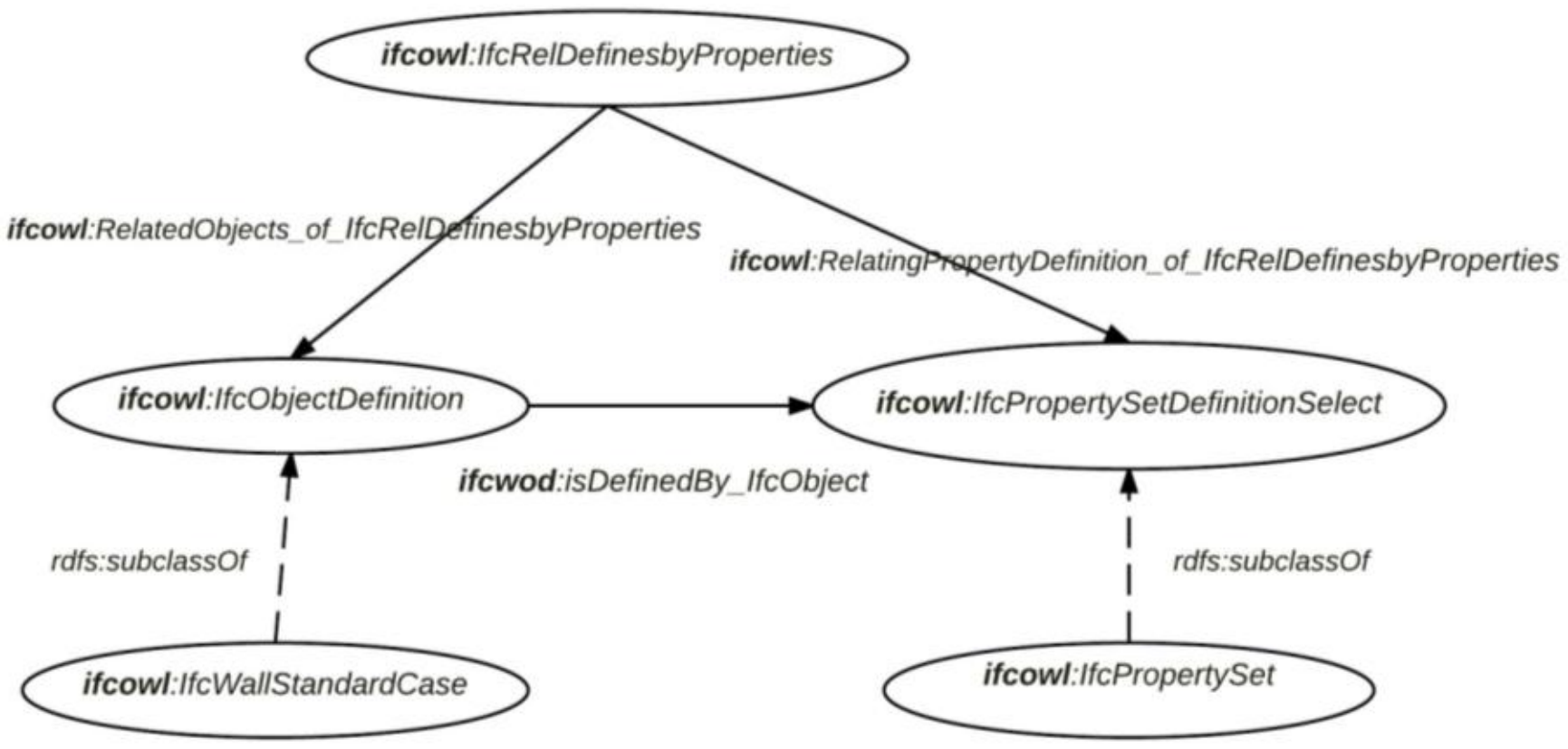

**Figure 2**An example of a subclass of IfcRelationship that relates the IfcWallStandardCase class with the IfcPropertySet class.

The subtypes of *IfcRelationship*havetwo attributes for relating IFC objects: Relating[name of relating object] and Related[name of related object]. Moreover, the IFC entity which models the related object has inverse attributes for referring to *IfcRelationship* entities. We take advantage of the inverse and Relating[name of relating object] attributesfor conceiving our OWL properties. Thus, we semantically adapt *IfcRelationship* entities into OWL language. For doing so, we parse the IFC standard specification in EXPRESS language(buildingSMART, 2015)to get these attribute specifications and conceive, partially, the IfcWoD ontology. For illustrating this procedure, let us consider the EXPRESS specification of *IfcRelSequence*and *IfcProcess*entities(see Figure 3).Our approach parses the inverse attributes of *IfcProcess*forcomposing the set:*I(IfcProcess)* = {(*isPredecessorTo*, *IfcProcess*), (*isSuccessorFrom*, *IfcProcess*), (*operatesOn*, *IfcProcessSelect*) }. For each tuple *(p, r)*in *I(e),* an OWL object property is created as follows: *ifcwod:p_erdf:typeowl:ObjectProperty*(e.g.:*ifcwod:isPredecessorTo_IfcProcessrdf:typeowl:ObjectProperty*).Itsdomain is the IfcOWL class that represents the IFC entity *e*using OWL (e.g.:*ifcwod:isPredecessorTo_IfcProcessrdfs:domainifcowl:IfcProcess*) and its rangeis the IfcOWL class that represents the IFC entity *r*(e.g.:*ifcwod:isPredecessorTo_IfcProcessrdfs:rangeifcowl:IfcProcess*). The annotation property*ifcwod:p_erdfs:label p^^xsd:string* is also asserted(e.g.:*ifcwod:isPredecessorTo_IfcProcessrdf:label "isPredecessorTo"^^xsd:string*).

```
ENTITY IfcRelSequence
 SUBTYPE OF (IfcRelConnects);
  RelatingProcess : IfcProcess;
  RelatedProcess : IfcProcess;
  TimeLag : OPTIONAL IfcLagTime;
  SequenceType : OPTIONAL IfcSequenceEnum;
  UserDefinedSequenceType : OPTIONAL IfcLabel;
 WHERE
  ...
END_ENTITY;
```

```
ENTITY IfcProcess
 ABSTRACT SUPERTYPE OF(ONEOF(IfcEvent, IfcProcedure, IfcTask))
 SUBTYPE OF (IfcObject);
  Identification : OPTIONAL IfcIdentifier;
  LongDescription : OPTIONAL IfcText;
 INVERSE
  IsPredecessorTo : SET OF IfcRelSequence FOR RelatingProcess;
  IsSuccessorFrom : SET OF IfcRelSequence FOR RelatedProcess;
  OperatesOn : SET OF IfcRelAssignsToProcess FOR RelatingProcess;
END_ENTITY;
```

**Figure 3**EXPRESS specification of *IfcRelSequence*and*IfcProcess* entities.

In the IFC m[illegible]as four attributes: *GlobalId*, *OwnerHistory*, *Name* and*Description*. Consequently, *IfcRelationship*inherits these four attributes. The *GlobalId* is a non-optional attribute as it allows to globally and uniquely identify a relationship within the whole software environment. These attributes are mapped as OWL object properties in IfcOWL. We consider these properties as part of a meta-model of our proposed IfcWoD ontology. Besides in the context of Web of data, a"*GlobalId*" property isredundantas a "universal" identifier already exists (i.e.: URI – Universal Resource Identifier)at a Web scale (not limited to a specific software). Nevertheless, *GlobalId* values can be a crucial information in the context of some IFC-based software (outside of Web). Because of this,datafrom attributes(*GlobalId*, *OwnerHistory*,*Name*, etc) aredescribed in IfcOWL and we ponder them as meta-data of various relations in IfcWoD. Inverse properties in IfcOWL are responsible for linking the relationship meta-data of severalIfcWoD properties. For illustrating this, let us consider one instance *i* of the*ifcowl:IfcProcess*class (Figure 3 illustrates the EXPRESS specification of *IfcProcess* entity). *i*asserts the *ifcowl*:*IsPredecessorTo*property (inverse of *ifcowl:RelatingProcess* in *IfcRelSequence*) and its value isthe instance *s* of *ifcowl:IfcRelSequence* class. The instance *s*encapsulates meta-data of the *ifcwod*:*isPredecessorTo_IfcProcess* propertythat defines the successor of the process *i,* namely*ifcowl:IfcProcess*.Meta-data of this property assertion comprises*GlobalId* (from the IFC software used to create this entity), *OwnerHistory*, *Name*, *Description, TimeLag, SequenceType*and*UserDefinedSequenceType.*These attributes are all defined as optional in the IFC standard, except for *GlobalId*. Therefore,we use the property *ifcwod*:*isPredecessorTo_IfcProcess* for capturing the main meaning of the*IfcRelSequence* entity.

We analysed all subtypes of *IfcRelationship*and we have identified relationship entities which have specific attributes defined within. In some cases, these specific attributes improve the semantics of a relationship, but in most cases these attributes are optional. For example, in the case of *IfcRelConnectsPathElements,*the attributes *ConnectionGeometry, RelatingPriorities*,*RelatedPriorities*, *RelatedConnectionType* and *RelatingConnectionType*semantically enrich the considered relationship. However, only *RelatedConnectionType* and *RelatingConnectionType*attributesare mandatory. When considering the IFC standard, there are 41 *IfcRelationship* subtypes that can be instantiated.Among them, only 14 relationship entities have been identified as havingadditional semantic attributes (besides Relating[name of relating object] and Related[name of related object] attributes which hold the main semantics of an IFC relationship). Therefore, the mapping of the majority *IfcRelationship*entitysubtypes as OWL object properties in IfcWoDcomprises the whole semantics of these relationships.

### 3.2 Adapting IfcPropertyAbstraction entity into OWL ontology

The entity *IfcPropertySet* is responsible for capturing properties common toseveral IFC objects. The naming convention "*Pset_Xxx*" is used for defining these property sets in the IFC standard.IFC users can extend these property setswith non-standard properties. IfcOWLrelies on the equivalent IFC schema since all IFC entities are mapped as OWL classes independently of their semantics. Moreover, “Property Sets” are not included in the consideredIfcOWL’sTBox. Consequently, several properties are instances of *ifcowl:IfcSimpleProperty*(subclass of *ifcowl:IfcProperty*)instead of being OWL properties in the ontology’s Tbox.Therefore, we propose to define these properties in our ontology schema, once again in order to respect the above mentioned ontology definition (e.g.: explicit specification). Besides in the Linked Data context, such modelling allows sharing vocabulary terms for publishing data on the Web. For illustrating a disadvantage of only using the IfcOWL vocabulary, let us consider a user that is searching a property for defining the gross planned area (e.g.: *:grossPlannedArea*) of a space.The*:grossPlannedArea*property does not exist in IfcOWL, but it is implicitly defined using the property set*Pset_SpaceCommon*from the IFC standard and an instance of *ifcowl:IfcSimpleProperty* class. Not only does the user has to browse the IfcOWL ontology for resource referencing and linking, but he/she must also investigate the IFC standard in order to find out that such property exists. Afterwards, he/she must manipulate unstructured information because IfcOWL conceives this property as a string value of the property *ifcowl:Name_of_IfcProperty*(e.g.: *:instance_of_IfcSimplePropertyifcowl:Name_of_IfcProperty “grossPlannedArea”xsd:string*). Thus, a simple typing error of this property name implies the impossibilityfor liking data. This could have been avoided if this property would have been formally defined in the ontology schema.

For semantically adapting the*IfcProperty* entity, we define the following IfcWoD properties, as listed in Table 1. *IfcProperty* entity has two subtypes: *IfcSimpleProperty*and*IfcComplexProperty*. They are mapped as OWL object properties *ifcwod:hasSimpleProperty* and *ifcwod:hasComplexProperty*, respectively. The *IfcSimpleProperty*subtypes are mapped assub-properties of *ifcwod:hasSimpleProperty* (see

Table 1).In addition, *ifcwod:hasComplexProperty*is asserted as an anti-reflexive property because an*IfcComplexProperty*[2] cannot reference itself.

**Table 1**Properties from IfcWoD for adapting *IfcProperty* into OWL object properties

| IfcWoD Property | Domain | Range | Sub-property of |
|---|---|---|---|
| *ifcwod:hasSimpleProperty* | *ifcowl:IfcPropertySet or ifcowl:IfcComplexPrope rty* | *ifcowl:IfcValue or ifcowl:ENUMERATION or ifcowl:IfcObjectReferenceS elect* | *owl:topObjectProperty* |
| *ifcwod:hasComplexProperty* | *ifcowl:IfcPropertySet or ifcowl:IfcComplexPrope rty* | *ifcowl:IfcComplexProperty* | *owl:topObjectProperty* |
| *ifcwod:hasReferenceValue* | - | *ifcowl:IfcObjectReferenceS elect* | *ifcwod:hasSimpleProper ty* |
| *ifcwod:hasSingleValue* | - | *ifcowl:IfcValue* | *ifcwod:hasSimpleProper ty* |
| *ifcwod:hasListValue* | - | *ifcowl:IfcValue* | *ifcwod:hasSimpleProper ty* |
| *ifcwod:hasEnumeratedValue* | - | *ifcowl:ENUMERATION* | *ifcwod:hasSimpleProper ty* |
| *ifcwod:hasTableValue* | - | *ifcowl:IfcValue* | *ifcwod:hasSimpleProper ty* |
| *ifcwod:hasBoundedValue* | - | *ifcowl:IfcValue* | *ifcwod:hasSimpleProper ty* |

In the IFC standard, there are more than 400 property sets defined for different IFC objects. They are also available as XML files by using the XSD schema*PSD_IFC4.xsd*(see*http://buildingSMART-tech.org/xml/psd/PSD_IFC4.xsd)*. Table 2contains the main correspondences for including these property sets in the IfcWoD'sTBox. This table's rows summarize the mapping between Property Set Definitions (PSD) using the *PSD_IFC4*XSD schema and OWL.

**Table 2**Properties from IfcWoD for adapting *IfcProperty* into OWL

| Mapping Rule | Property Set Definition (PSD) XSD | OWL language |
|---|---|---|
| **R1** | *<xs:complexType name="PropertyDef">* | *owl:ObjectProperty* |
| **R2** | *<xs:element type="xs:string" name="Name">* | *rdfs:label and URN* |
| **R3** | *<xs:element type="xs:string" name="Definition">* | *rdfs:comment* |
| **R4** | *<xs:element name="NameAliases">* | *rdfs:label (@lang)* |
| **R5** | *<xs:element name="DefinitionAliases">* | *rdfs:comment (@lang)* |
| **R6** | *<xs:element type="PropertyType" name="PropertyType">* | *rdfs:subPropertyOf and rdfs:range* |

Our approach parses those XML files containing properties information for conceiving OWL object properties. EveryPSD XML file contains meta-data of one property set and meta-data for its properties. The property set name is used to compose the URI of properties in this set.

[2]www.buildingsmart-tech.org/ifc/review/IFC4Add1/rc1/html/schema/ifcpropertyresource/lexical/ifccomplexproperty.htm

For exemplifying this process, let us consider the PSD *Pset_StackTerminalTypeCommon*[3]. A portion of this PSD is illustrated in Figure 4. The property set nameis retrieved from the element <*Name*> that has text content. We parse this text content for creating the namespace *http://buildingsmart.org/ontology/ifcwod/Pset_StackTerminalTypeCommon#*andits prefix*pset_StackTerminalTypeCommon*. Thus, we define a namespace per property set, which guarantees a unique URI for each property (i.e. resource) in this set. Afterwards, other elements are parsed and mapped following the main rules listed in Table 2. RuleR1 is appliedfor each <*PropertyDef*>XML element and an OWLobjectproperty is created. Its universal resource name (URN) is determined by applying rule R2 (e.g. *pset_StackTerminalTypeCommon:referencerdf:typeowl:ObjectProperty* ). Besides, a *rdfs:label* annotation property is asserted for the so-created property by applying rule R2 (e.g.: *pset_StackTerminalTypeCommon:referencerdfs:label “Reference”^^xsd:string*). By applying rules R3 and R4 to the contents of thefirst<*PropertyDef*> elementin Figure 4, the following triples are asserted:*pset_StackTerminalTypeCommon:referencerdfs:comment“Reference ID for this specified…”^^xsd:string;pset_StackTerminalTypeCommon:referencerdfs:label “Reference”@en;pset_StackTerminalTypeCommon:referencerdfs:label “参照記号”@ja-Jpan*.The mapping rule R5 is applied to <*DefinitionAliases*> contents that are not detailed in Figure 4. R6 maps the <*PropertyType*> element for the previous property definition example as follows: *pset_StackTerminalTypeCommon:referencerdfs:subPropertyOfifcwod:hasSingleValue* and

```
- <PropertySetDef xsi:noNamespaceSchemaLocation="http://buildingSMART-tech.org/xml/psd/PSD_IFC4.xsd"
  templatetype="PSET_TYPEDRIVENOVERRIDE" ifdguid="ad7e4680d20c11e1800000215ad4efdf"
  xmlns:xsd="http://www.w3.org/2001/XMLSchema" xmlns:xsi="http://www.w3.org/2001/XMLSchema-instance">
    <IfcVersion version="IFC4"/>
    <Name>Pset_StackTerminalTypeCommon</Name>
    <Definition>Common properties for stack terminals.</Definition>
    <Applicability/>
  + <ApplicableClasses>
    <ApplicableTypeValue>IfcStackTerminal</ApplicableTypeValue>
  - <PropertyDefs>
    - <PropertyDef ifdguid="b2db9100d20c11e1800000215ad4efdf">
        <Name>Reference</Name>
        <Definition>Reference ID for this specified type in this project (e.g. type 'A-1'), provided, if there is no classification
          reference to a recognized classification system used.</Definition>
      - <PropertyType>
        - <TypePropertySingleValue>
            <DataType type="IfcIdentifier"/>
          </TypePropertySingleValue>
        </PropertyType>
      - <NameAliases>
          <NameAlias lang="en">Reference</NameAlias>
          <NameAlias lang="ja-JP">参照記号</NameAlias>
        </NameAliases>
      + <DefinitionAliases>
      </PropertyDef>
    - <PropertyDef ifdguid="b838db80d20c11e1800000215ad4efdf">
        <Name>Status</Name>
        <Definition>Status of the element, predominately used in renovation or retrofitting projects. The status can be
          assigned to as "New" - element designed as new addition, "Existing" - element exists and remains, "Demolish"
          element existed but is to be demolished, "Temporary" - element will exists only temporary (like a temporary
          support structure).</Definition>
      - <PropertyType>
        - <TypePropertyEnumeratedValue>
          - <EnumList name="PEnum_ElementStatus">
              <EnumItem>NEW</EnumItem>
              <EnumItem>EXISTING</EnumItem>
              <EnumItem>DEMOLISH</EnumItem>
              <EnumItem>TEMPORARY</EnumItem>
```

**Figure 4**A portion of the PSD *Pset_StackTerminalTypeCommon*

*pset_StackTerminalTypeCommon:referencerdfs:rangeifcowl:IfcIdentifier*. *reference*is defined as a sub-property of*ifcwod:hasSingleValue* because the <*PropertyType*> element contains a <*TypePropertySingleValue*> element.Moreover, the *type* attributefrom the<*DataType*> element defines the property range.

All mapping rules from Table 2are also applied to the second <*PropertyDef*> element in the PSD illustrated in Figure 4. Hence, *pset_StackTerminalTypeCommon:status*is a sub-property of*ifcwod:hasEnumeratedValue* and the <*EnumList*> element (e.g.: PEnum_ElementStatus) is mapped as subclass of *ifcowl:ENUMERATION*(e.g.:

[3] http://www.buildingsmart-tech.org/ifc/IFC4/Add1/html/psd/Pset_StackTerminalTypeCommon.xml

*pset_StackTerminalTypeCommon*:*PEnum_ElementStatusrdfs:subClassOfifcowl:ENUMERATION*). Element <*EnumList*> (i.e.: <*EnumItem*>) contains instances of *pset_StackTerminalTypeCommon:PEnum_ElementStatus*. ThroughR6, we defined the range for the property *status (e.g.: pset_StackTerminalTypeCommon:statusrdfs:rangepset_StackTerminalTypeCommon:PEnum_ElementStatus).* Therefore, we semantically adapt the*IfcPropertyEnumeration*entity into OWL*.*

All IFC entities that are subtypes of *IfcSimpleProperty*(except *IfcPropertyReferenceValue* and *IfcEnumeratedValue*) have a "Unit" attribute for asserting the unit used for expressing property values. For describing this information within IfcWoD, we define the property *ifcwod:hasUnit,*having*ifcowl:IfcUnit* as its range and *ifcowl:IfcValue*as its domain.Thus, the unit of a property value is described as part of value semantics instead of property semantics. By doing this, we do not have to define in the Tbox one property perconsidered unit value.

Due to limited page number, in this article we focus solely on the semantical adaptation into OWL of the *IfcPropertyAbstraction* subtypes: *IfcProperty* and *IfcPropertyEnumeration.* The othersubtypes of*IfcPropertyAbstraction*(i.e.*IfcExtendedProperties* and *IfcPreDefinedProperties)* are not discussed here.

## 4 Results and discussions

IfcWoD allows enhancing reasoning (e.g.: data inference) over building data because it allows taking advantage of various OWL built-in classes.Notablyit allows defining logical characteristics for the considered properties (e.g.: *owl:TransitiveProperty, owl:SymmetricProperty, owl:ReflexiveProperty*). For example, the property*ifcwod:isPredecessorTo_IfcProcess*for the*ifcowl:IfcProcess* class can be specified as a transitive property. This specification increases data inference when applying a description logic basedreasoner. This is not possible using exclusively the IfcOWL vocabulary because the considered relationship is mapped as an OWL class.

Another advantage of our approach is the fact that it allows easier query writing and relationship understanding. This is because we do not have to browse*ifcowl:IfcRelationship* instances for relating IFC objects. For illustrating this, let us suppose the following SPARQL[4](query language for RDF) query, built only using IfcOWL terms: *SELECT ?x ?z { ?x ifcowl:IsPredecessorTo ?y. ?yifcowl:RelatingProcess_of_IfcRelSequence ?z. }*. If we use IfcWoD terms, this query is simplified to *SELECT ?x ?y {?x ifcwod:isPredecessorTo_IfcProcess ?y}*. Moreover, if *ifcwod:isPredecessorTo_IfcProcess* is stated as a transitive property more information can be inferred, for example, if P1 is predecessor to P2 and P2 is predecessor to P3, the OWL reasoner infers that P1 is also predecessor to P3 where P1, P2 and P3 are instances of *ifcowl:IfcProcess* class.

Adapting the subtypes of the*IfcPropertyAbstraction*entity into OWL (as described in subsection3.2) has four main advantages: (a) it simplifies query writing; (b) it improves query response time; (c) it allows sharing building properties in the linked data context and (d) it reduces data redundancy. Table 3illustrates the advantage (a), by listing several SPARQL queries rewritten using IfcWOD terms. When comparing those queries, namely (Q1, Q1'), (Q2, Q2') and (Q3, Q3'), we may notice that the amount of triples is reducedof about 50%.Qn is a SPARQL query without IfcWoD terms and Qn' is a query with IfcWoD vocabulary where n $\in$ {1, 2, 3}.Q1 and Q1' retrieve all external walls of a building project. Q2 and Q2' retrieve all doors and their references. Q3 and Q3' retrieve spaces which are above an internal reference height of the building and their corresponding references.

Table 4 presents response time benchmark resultsfor the above considered queries. For these experiments, we have used a 3.0.1 Stardog[5] triple store which played the role of the server and was encapsulated in a virtual machine with the following configuration: one microprocessor Intel Xeon CPU E5-2430 at 2.2GHz with 2 cores out of 6, 8GB of DDR3

[4] http://www.w3.org/TR/rdf-sparql-query/
[5]http://docs.stardog.com/

RAM memory and the “Java Heap” size for the Java Virtual Machine set to 6GB. We populated the IfcOWL and IfcWoD ontologies with building information from an IFC-STEP file of 11.5 MB size. This file was mapped into more than one million RDF triples and stocked jointly with IfcOWL and IfcWoD over one knowledge base in the triple store. The client machine has the following configuration: one microprocessor Intel Core CPU I7-4790 at 3.6GHz with 4 cores, 8GB of DDR3 RAM memory at 1600MHz and the “Java Heap” size set to 1GB.

Table 4 shows means and standard deviations of 30 executions for each query as requested by the client machine. We can conclude that the query mean time execution was reduced of about 90% or 95% in experiments which use IfcWoD terms for querying. Besides, executed queries retrieve the same results by using or not IfcWoD ontology as expected. This allows justifying advantage (b).

**Table 3**Comparison between queries with and without IfcWoD ontology

| | Querying solely with IfcOWL terms | | Querying with IfcWoD terms |
|---|---|---|---|
| **Q1:** | *SELECT ?wall WHERE {<br>?wall rdf:typeifcowl:IfcWall;<br>ifcowl:IsDefinedBy_of_IfcObject ?rel.<br>?relifcowl:RelatingPropertyDefinition ?pSet.<br>?pSetifcowl:HasProperties_of_IfcPropertyS<br>et ?p.<br>?p<br>rdf:typeifcowl:IfcPropertySingleValue;<br>ifcowl:Name_of_IfcProperty ?name.<br>?name ifcowl:has_string<br>"IsExternal"^^xsd:string.<br>?p ifcowl:NominalValue ?val.<br>?valifcowl:has_boolean<br>"true"^^xsd:boolean. }* | **Q1’:** | *SELECT ?wall WHERE {<br>?wall rdf:typeifcowl:IfcWall;<br>**ifcwod**:isDefinedBy_IfcObject ?pSet.<br>?pSet**pset_WallCommon**:isExternal ?val.<br>?valifcowl:has_boolean<br>"true"^^xsd:boolean. }* |
| **Q2:** | *SELECT ?door?reference WHERE {<br>?doorrdf:typeifcowl:IfcDoor;<br>ifcowl:IsDefinedBy_of_IfcObject ?rel.<br>?relifcowl:RelatingPropertyDefinition ?pSet.<br>?pSetifcowl:HasProperties_of_IfcPropertyS<br>et ?p.<br>?p<br>rdf:typeifcowl:IfcPropertySingleValue;<br>ifcowl:Name_of_IfcProperty ?name.<br>?name ifcowl:has_string<br>"Reference"^^xsd:string.<br>?p ifcowl:NominalValue ?val.<br>?valifcowl:has_string?reference. }* | **Q2’:** | *SELECT ?door?reference WHERE {<br>?doorrdf:typeifcowl:IfcDoor;<br>**ifcwod**:isDefinedBy_IfcObject ?pSet.<br>?pSet**pset_DoorCommon**:reference ?val.<br>?valifcowl:has_string?reference. }* |
| **Q3:** | *SELECT ?x ?reference WHERE {<br>?floor rdf:typeifcowl:IfcBuildingStorey.<br>?floor ifcowl:Elevation ?elev.<br>?elevifcowl:has_double ?y.<br>?floor ifcowl:isDecomposedBy ?rel.<br><br>?relifcowl:RelatedObjects_of_IfcRelAggregates<br>?x.<br>?x rdf:typeifcowl:IfcSpace;<br>ifcowl:IsDefinedBy_of_IfcObject ?rel.<br>?relifcowl:RelatingPropertyDefinition ?pSet.<br>?pSetifcowl:HasProperties_of_IfcPropertyS<br>et ?p.<br>?p<br>rdf:typeifcowl:IfcPropertySingleValue;<br>ifcowl:Name_of_IfcProperty ?name.<br>?name ifcowl:has_string<br>"Reference"^^xsd:string.<br>?p ifcowl:NominalValue ?val.<br>?valifcowl:has_string?reference.<br>FILTER (?y > 0) }* | **Q3’:** | *SELECT ?x ?reference WHERE {<br>?flrdf:typeifcowl:IfcBuidingStorey.<br>?flifcowl:Elevation ?elev.<br>?elevifcowl:has_double ?y.<br><br>?fl**ifcwod**:isDecomposedBy_IfcObjectDefinition<br>?x.<br>?x rdf:typeifcowl:IfcSpace;<br>**ifcwod**:IsDefinedBy_IfcObject?pSet.<br>?pSet**pset_SpaceCommon**:reference ?val.<br>?valifcowl:has_string?reference.<br>FILTER (?y > 0) }* |

**Table 4**Analysis of the query performance

| | Q1 | Q1’ | Q2 | Q2’ | Q3 | Q3’ |
|---|---|---|---|---|---|---|
| **Mean(seconds)** | 0.242 | **0.026** | 0.516 | **0.025** | 1.348 | **0.056** |
| **Standard Deviation** | 0.024 | **0.009** | 0.019 | **0.008** | 0.024 | **0.017** |
| **#Results** | **37** | **37** | **141** | **141** | **67** | **67** |
| **Mean Time Reduction (%)** | | **89.26%** | | **95.15%** | | **95.85%** |

Advantage (c) is justified by the fact that the *IfcProperty* entity is formally and explicitly defined in ourTBox.When consideringonly the IfcOWL ontology, properties are instances of the*IfcProperty* class and they cannot be easily shared in the Linked Datacontext, asthey are not part of theIfcOWL vocabulary.

For justifying (d), let us consider the property “*IsExternal*” presentin the*Pset_WallCommon*IFC property set. Figure 5illustrates how “*IsExtenal*” data is described without and with IfcWoD vocabulary. When considering solely the IfcOWL ontology, for relating this property to an *IfcPropertySet* instance one has to instantiate the*IfcPropertySingleValue* class. Thus, describing this property withoutIfcWoD implies replicatingthe same assertions(e.g.: *:PropertyInstanceifcowl:Name_of_IfcProperty“IsExtenal”^^xsd:string*)for each instance of the*IfcPropertySet*that should contain an “*IsExternal*” property.Such data replication is avoided when using the IfcWoDproperty: *pset_WallCommon:isExternal*.

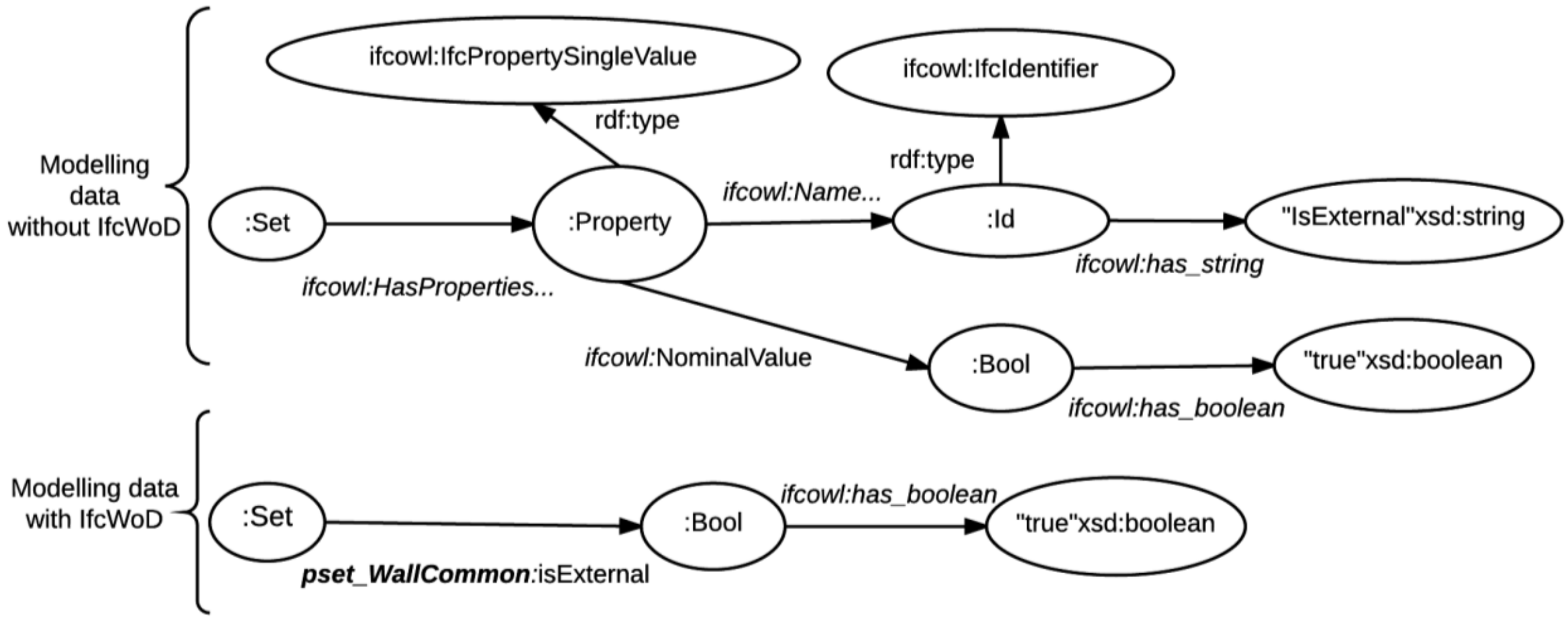

**Figure 5**Modelling “IsExternal” property from *Pset_WallCommon* without and with IfcWoD.

## 5 Conclusion

In this work, we proposed an approach that focuses onsemantic modelling IFC relations in OWL. For doing so, we have based our proposal on widely accepted guidelines for ontology definition and ontology modelling. We described a semi-automatic method for conceiving an ontology (i.e. IfcWoD) more adequate for semantically linking IFC data in thecontext of the Web of Data. We have rigorouslyanalysed modelling choices as applied in the current state-of-the-art IfcOWL ontology. Based on the disadvantages identified, we propose a novel modelling that allowsan eased application of the Linked (Open) Data principles. Our IfcWoD ontology is not another version of IfcOWL, but a new ontology that uses terms from IfcOWL. Moreover, a part of IfcOWL ontology can be considered as a meta-model forIfcWoD. The experiment results and discussions prove that IfcWoD linked to IfcOWL simplifies querying writing and improves query response time for retrieving building data, when compared to only consideringIfcOWLontology. Additionally, defining PSD properties in the ontology’sTBox allows reducing data redundancy.

Further workaddressesanalysing the trade-off between data redundancy and query performance, when defining the domain of PSD properties directly as IFC object classes instead of*IfcPropertySet* classes. Moreover, there are many properties such as "*reference*" that are present in several PSDs. Thus, we want to further study the gathering and the hierarchizationof such common properties.

## Acknowledgements

This work has been financed by the French company ACTIVe3D (see http://www.active3d.net/fr/) and supported by the Burgundy Regional Council (see http://www.region-bourgogne.fr/).